\documentclass[final]{cvpr}

\usepackage{times}
\usepackage{epsfig}
\usepackage{graphicx}
\graphicspath{{images/}}
\usepackage{amsmath}
\usepackage{amssymb}

\usepackage{comment}
\usepackage{color}
\usepackage{wrapfig}
\usepackage{tabularx}
\newcolumntype{C}{>{\centering\arraybackslash}X}
\usepackage{stfloats} 
\usepackage[table]{xcolor}
\usepackage{multicol} 
\usepackage{lipsum} 

\usepackage[pagebackref=true,breaklinks=true,colorlinks,bookmarks=false]{hyperref}

\def\cvprPaperID{15} 
\def\confYear{CVPR 2021}

\begin{document}

\title{Towards Efficient Convolutional Network Models \\ with Filter Distribution Templates}

\author{Ramon Izquierdo-Cordova, Walterio Mayol-Cuevas\\
Department of Computer Science, University of Bristol, United Kingdom\\
\url{https://vilab.blogs.bristol.ac.uk}\\
{\tt\small \{ramon.izquierdocordova,walterio.mayol-cuevas\}@bristol.ac.uk}
}

\def\AcknowledgmentsText{This work was partially supported by CONACYT and the Secretar\'ia de Educaci\'on P\'ublica, M\'exico.}


\maketitle

\begin{abstract}

Increasing number of filters in deeper layers when feature maps are decreased is a widely adopted pattern in convolutional network design. It can be found in classical CNN architectures and in automatic discovered models. Even CNS methods commonly explore a selection of multipliers derived from this pyramidal pattern. We defy this practice by introducing a small set of templates consisting of easy to implement, intuitive and aggressive variations of the original pyramidal distribution of filters in VGG and ResNet architectures. Experiments on CIFAR, CINIC10 and TinyImagenet datasets show that models produced by our templates, are more efficient in terms of fewer parameters and memory needs.

\end{abstract}

\section{Introduction}

Despite the continuous progress in convolutional neural network (CNN) models, there has been an element in their design remaining unchanged. There is a practice of increasing the number of filters in deeper layers basically doubling the filters when a pooling layer halves the resolution of the feature map. It is generally believed that a progressive increase in the number of kernels compensates a possible loss of the representation caused by the spatial resolution reduction \cite{lecun1998gradient}, as well as it improves performance by keeping a constant number of operations in each layer \cite{chu2014analysis}.

This pattern was first proposed in \cite{lecun1998gradient} with the introduction of LeNet and can be observed in a diverse set of models such as VGG\cite{simonyan2014very}, ResNet\cite{he2016deep} and MobileNet\cite{howard2017mobilenets}. Even models obtained from neural architecture search (NAS), such as NASNet \cite{zoph2017learning}, follow this principle since automatic model discovery methods are mainly formulated to search for layers and connections while the number of filters in each layer remains fixed. 

\begin{figure}
    \begin{center}
        \includegraphics[width=0.9\linewidth]{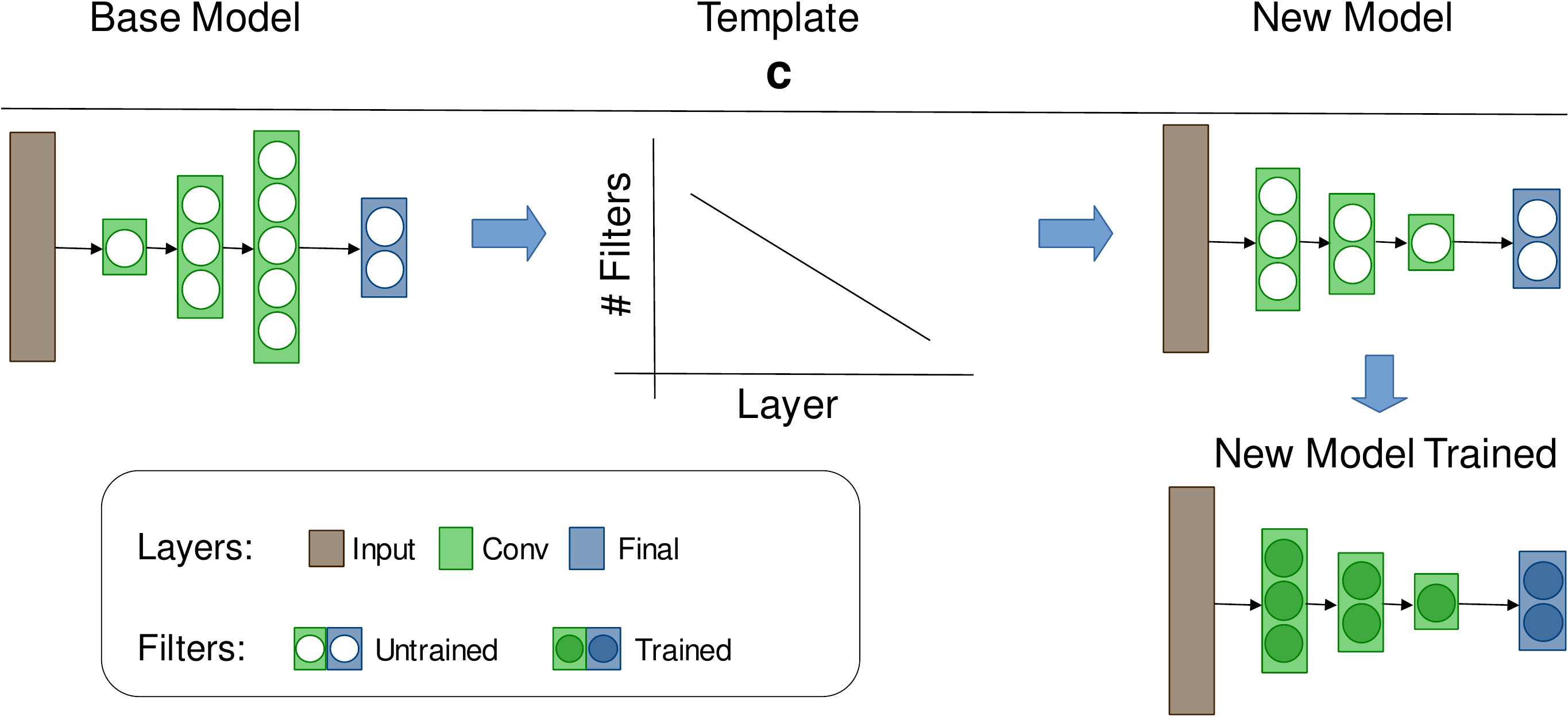}
    \end{center}
    \caption{Increasing filters per layer is a common design in convolutional models (left). We propose to apply a small set of linear templates to an existing model to change its filter distribution (center). After being trained from scratch (right), resulting models are competitive in accuracy compared to the original model, however, they require less computational resources.}
\label{fig:templates_training}
\end{figure}

Techniques used by NAS are now applied to channel number search (CNS) to find the optimal distribution of filters in a convolutional neural network \cite{wang2020revisiting} based on the intuition that the incremental design could not be the best option. However, approaches reduce the search process considering a limited set of promising values that are multiples of a base pattern, normally the aforementioned pyramidal distribution. Another limitation for widely using CNS algorithms resides in the high computational cost implied in the exploration of the search space which requires training and evaluating a vast number of candidate models before finding a good distribution of filters.

To overcome this limitation we build on the idea that, instead of exploring for individual widths in each layer, we should explore a few simple and radically different distributions. In this work we introduce a small set of these predefined distributions, called templates, intuitively created by using simple linear equations or combinations of them that can be easily implemented in most of the existing CNN classical models.

As depicted in figure \ref{fig:templates_training}, to use a template, we take a base model and redefine its filter distribution with the pattern provided. Once trained, the new model requires similar FLOPs and produces better or at least competitive accuracy, but consumes less of other resources. We haven't defined a way of finding the most appropriated template. Being the set of templates small enough, a simple sequential search is still fairly competitive in time compared to automatic methods. However, experiments show an emerging pattern on which template presents favourable features according to the requirements of the task.

Experimental evidence shows that simple changes to the pyramidal distribution of filters in CNN models lead to improvements in accuracy, number of parameters or memory footprint; we highlight that tested models, although significantly changed in their original filter design, present high resiliency in accuracy, a phenomena that requires further research and explanation.\\

\section{Related Work}\label{chap:related}

Classical model design has been performed by extensive and computationally demanding experimentation, relying strongly on designers' experience. Widely employed architectures such as VGG \cite{simonyan2014very}, ResNet \cite{he2016deep}, Inception \cite{szegedy2015going} and MobileNet \cite{howard2017mobilenets} have been build on heuristics known to work well. Recently, methods for automatic model discovering have reached state-of-the-art accuracy at the cost of exploring vast number of models with different layer types and connections \cite{zoph2017learning,zoph2018learning,liu2018darts,tan2019mnasnet,ren2020comprehensive}.

It is implicitly assumed in pruning methods that the original distribution of filters is not the optimal for CNN architectures. Although the goal in the field is to achieve a reduction of unnecessary parameters in a deep network model \cite{blalock2020state}, the process is frequently carried out by removing filters \cite{he2017channel,luo2017thinet,leclerc2018smallify,he2019filter,you2019gate} using some heuristic. Thus, the final model ends up with a different distribution of filters.

New methods for channel number search (CNS) aim to automatically find the best number of filters for each layer in a neural network \cite{gordon2018morphnet,lee2020neuralscale}, usually reducing the computational burden, caused for the exploration of the search space, by parameter sharing \cite{yu2019autoslim,dong2019network,wang2020revisiting,berman2020aows}. However, most of the hand-crafted architectures, and those used as base models to initialise the automatic search in NAS and CNS methods, share the practice of increasing filters resembling LeNet design\cite{lecun1998gradient}.

We build on the findings of \cite{liu2018rethinking,frankle2020pruning} suggesting that accuracy obtained by pruning techniques can be reached by removing filters at initialisation to fit a certain resource budget and then training from scratch. Our work is related to \cite{zhu2020efficient} in the sense that we are also applying a reduced set of transformations to the original network. Our approach differs however, because it simply requires manually changing the distribution of filters in the base model according to the proposed templates and then, training the resulting model from scratch.

\section{Filter Distribution Templates}\label{chap:methods}

\begin{figure}
    \begin{center}
    \includegraphics[width=1.0\linewidth]{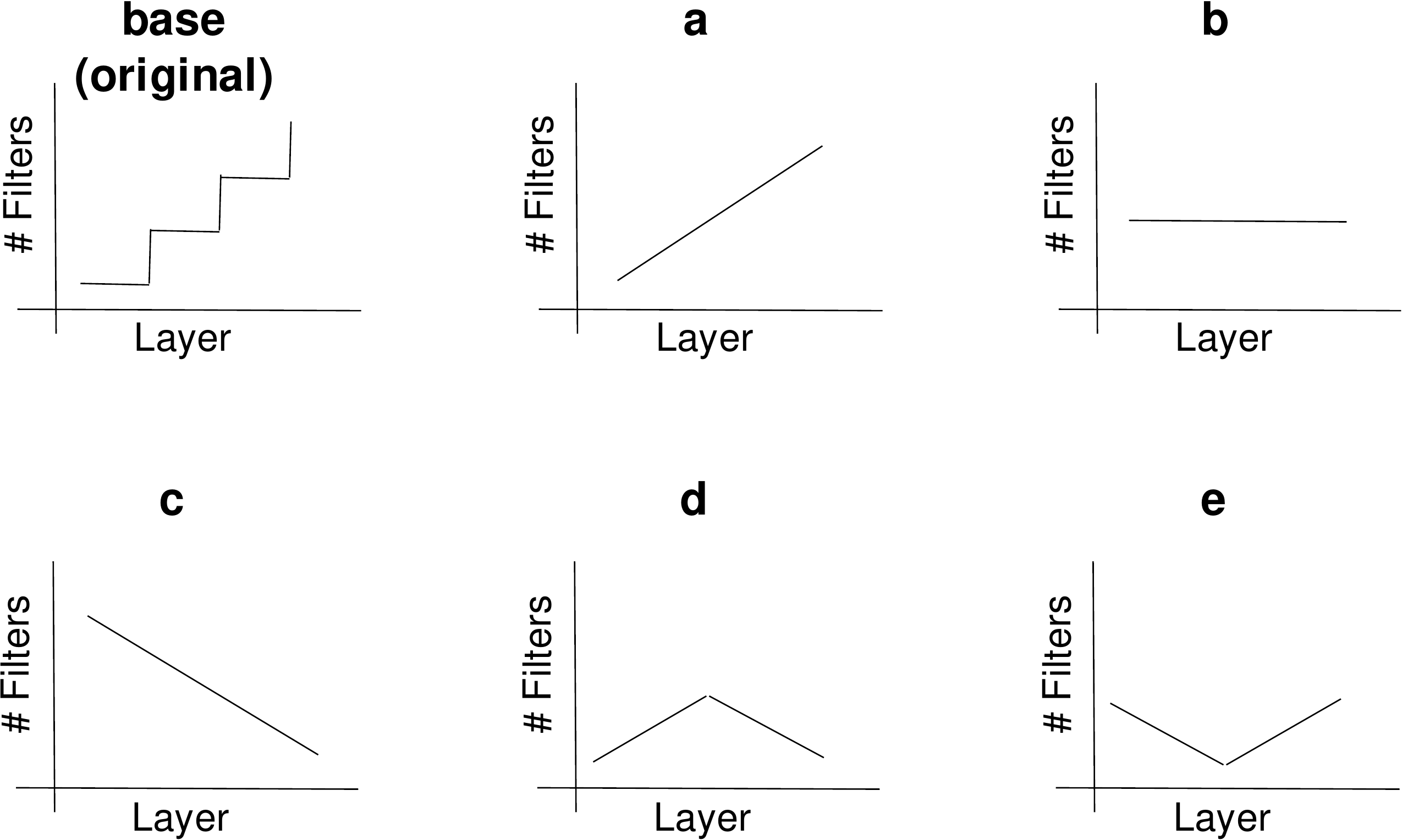}
    \end{center}
    \caption{Schematic distribution of filters per layer in linear templates. These templates are created with simple and intuitive but aggressive variations. Base distribution, which is the original distribution, shows the common design of growing the filters when resolution of feature maps decreases in deeper layers. We match the same number of filters in the thinnest layers and adjust the maximum to meet the original number of FLOPS for a fair comparison. }
\label{fig:templates}
\end{figure}

In this work, rather than attempting to find the optimal filter distribution with expensive automatic techniques, we propose to first adjust the filters of a convolutional network model via a small number of predefined templates. These templates are depicted in figure~\ref{fig:templates}, and have been found to perform well and are thus candidates for model performance improvement beyond accuracy. Performance criteria such as parameters, memory footprint and inference time are arguably as important.

We adopt as the first template, one with the same incremental distribution but with a smooth step \textbf{(a)}. The second template is a distribution with a fixed number of filters \textbf{(b)} as in the original Neocognitron design. Another immediate option, contrary to the increasing distribution, is a decreasing distribution of filters \textbf{(c)}. Inspired by the distributions of blocks from the resulting ResNet101 and VGG models found in \cite{gordon2018morphnet} and \cite{leclerc2018smallify,you2019gate}, we define a template in which filters agglomerate in the centre \textbf{(d)} and, on the contrary, where filters are reduced in the centre of the model \textbf{(e)}.

A first approach to implement a change of filters in a model is to keep the original number of filters in the resulting model and to redistribute them differently across its layers. But final models end up with different resource demands and therefore making a fair comparison is hard. Parameters, FLOPs and inference time have been used as a proxy for comparing models with different designs. We believed that using one metric is insufficient for a fair comparison. As a example, our implementations of VGG and ResNet count approximate number of parameters (20.03 millions versus 23.52 millions, respectively) but ResNet's FLOPs (1307 Mflops) are more than three times VGG ones (399 Mflops). To facilitate comparisons, we match one metric while comparing other. In particular, we fix models obtained from templates to match the number of FLOPs of the original distribution and then compare a second metric such as parameters, memory footprint or inference time.

\begin{table*}[ht!]
\centering
\caption{VGG19 performance with the original distribution of filters and five templates evaluated on cifar10, cifar100 and tinyImagenet datasets. Flops are kept to similar values (399 Mflops). After filter redistribution, most models surpass the base accuracy with less parameters. Results show average of three repetitions.\\}
\footnotesize
\begin{tabular}{l|rr|rr|rr|llll}
         & \multicolumn{2}{l|}{Param}      & \multicolumn{2}{l|}{Mem}  & \multicolumn{2}{l|}{Inference} &         &          &         & tiny     \\
Template & \multicolumn{2}{l|}{(Millions)} & \multicolumn{2}{l|}{(GB)} & \multicolumn{2}{l|}{Time (ms)} & cifar10 & cifar100 & cinic10 & Imagenet \\ \hline

base     & 20.03           & \% ↓          & 87.0         & \% ↓       & 1.80           & \% ↓          & 94.90 ± 0.10 & 73.91 ± 0.08 & 85.79 ± 0.10 & 57.34 ± 0.30 \\ \hline
a        & 17.23           & 13.9          & 76.5         & 12.0       & 1.68           & 6.6           & 95.03 ± 0.26 & 74.47 ± 0.10 & 86.13 ± 0.14 & 57.21 ± 0.56 \\
b        & 3.17            & 84.1          & 23.0         & 73.5       & 1.31           & 27.2          & 95.01 ± 0.10 & 73.34 ± 0.24 & 86.37 ± 0.02 & 56.88 ± 0.31 \\
c        & 1.89            & 90.5          & 17.8         & 79.5       & 1.35           & 25.0          & 95.04 ± 0.19 & 72.27 ± 0.35 & 86.15 ± 0.09 & 54.50 ± 0.24 \\
d        & 8.07            & 59.7          & 39.8         & 54.2       & 1.37           & 23.8          & 95.21 ± 0.08 & 74.64 ± 0.13 & 86.49 ± 0.05 & 59.45 ± 0.18 \\
e        & 2.06            & 89.7          & 17.8         & 79.5       & 1.39           & 22.7          & 94.75 ± 0.07 & 71.13 ± 0.28 & 85.85 ± 0.03 & 54.26 ± 0.35
\end{tabular}
\label{tab:templates_accuracy_vgg19}
\end{table*}

In a more formal way, we define a convolutional neural network base model as a set of numbered layers $\textsl{L}=1,... ,D+1$, each with $f_{l}$ filters in layer $l$. $D+1$ is the final classification layer which size is given by the task. The ordered set of all filters in the model is $F_{1:D} = \{ f_{1}, \dotsc, f_{D} \}$ and the total number of FLOPs, the resource to be matched between templates, is given by some function $\mathcal{R} ( F_{1:D} )$. We want to find a new distribution of filters $F'_{1:D}$ in which 
\begin{equation}  \label{flops_equialence}
    \mathcal{R} ( F_{1:D} ) \approx \mathcal{R} ( F'_{1:D} )
\end{equation}
and to test if the common heuristic of distributing $F_{1:D}$ having $f_{l+1} = 2f_{l}$ each time the feature map is halved, is advantageous to the model over $F'_{1:D}$ when evaluating performance, memory footprint and inference time.

Our templates are defined as simple linear segments, or a combination of them, in which $min(F'_{1:D}) = min(F_{1:D})$ and $max(F'_{1:D}) = n \in \mathbb{N}$ satisfying constrain (\ref{flops_equialence}).

\section{Experiments}\label{chap:experiments}

We investigate the effects of applying different templates to the distribution of kernels in well known convolutional neural network models (VGG and ResNet). We comparing highlight that resulting models obtained from templates have similar FLOPs than the original model. This constrain facilitates further comparison of models under the basis of size, memory and speed tested in four well known datasets for classification tasks.\\

\subsection*{Datasets and Models}

We selected four datasets with diverse number of samples and classes to test our templates but allowing a relatively fast training process. Each model needed to be evaluated with a set of five templates plus the original model, so we decided to use CIFAR-10, CIFAR-100 \cite{krizhevsky2009learning}, CINIC-10 \cite{darlow2018cinic} and Tiny-Imagenet \cite{le2015tiny}. The first two datasets contain sets of 50,000 and 10,000 colour images for train and validation respectively, with a resolution of 32x32. CINIC-10 contains 90,000 images in each, the training and validation sets with the same resolution and classes as the CIFAR-10 dataset. Tiny-Imagenet is a reduced version of the original Imagenet dataset with only 200 classes and images with a resolution of 64 x 64 pixels.

We evaluated VGG\cite{simonyan2014very} and ResNet\cite{he2016deep} models, which represent some of the most influential CNN architectures on the ImageNet challenge in previous years \cite{cocskun2017overview,russakovsky2015imagenet}.

\subsection*{Implementation Details}

Experiments have models fed with images with the common augmentation techniques of padding, random cropping and horizontal flipping and additionally, with cutout \cite{devries2017improved} using 1 patch of 16 x 16 pixels. Our experiments were run in a NVidia Titan X Pascal 12GB GPU adjusting the batch size to 64 for TinyImagenet and 128 for the rest of the datasets.
For CIFAR10, CIFAR100 and CINIC10, all models were trained for 200 epochs using the same conditions: stochastic gradient descent (SGD) with scheduled learning rate of 0.1 decreased with gamma 0.2 at epochs 60, 120 and 160; weight decay of 1e-5 and momentum of 0.9. For TinyImagenet, models were trained for 90 epochs using SGD with scheduled learning rate of 0.1 decreased with gamma 0.1 at epochs 45, 70 and 85; weight decay of 1e-1 and momentum of 0.9.

\subsection*{Effects of Templates on Base Models}

We conducted an experiment to test our proposed templates on the selected architectures. Tables \ref{tab:templates_accuracy_vgg19} and \ref{tab:templates_accuracy_resnet50} show properties of resulting models after using templates on VGG and ResNet respectively. Parameters, memory footprint and inference time are reduced in all the cases. This result is in some way surprising given that template patterns were only selected following simplicity and diversity but not precisely efficiency.

Particularly, in table \ref{tab:templates_accuracy_vgg19}, we observe increase in accuracy up to 2.11 points over the base model mostly obtained with template \textbf{d}. Remarkable reductions of 90\% in parameters, 79\% in memory usage and 25\% in inference time are produced by using template \textbf{c} while accuracy is still slightly superior on CIFAR10, CIFAR100 and CINIC10 datasets. The fastest model with a reduction of 28\% in inference time is reached wuth template \textbf{b}.

Behaviour for ResNet differs from that of VGG in some aspects. Impact in resource consumption es lower. Template \textbf{c} shows savings of 85\% in parameters, 30\% in memory usage and 20\% in inference time being the fastest of all. Highest accuracy is obtained by templates \textbf{d}, \textbf{b} and \textbf{a} on CIFAR, CINIC and Tiny-Imagenet respectively. The smallest model in memory is given by template \textbf{d}.

\begin{table*}
\centering
\caption{ResNet50 performance with the original distribution of filters and five templates evaluated on cifar10, cifar100 and tinyImagenet datasets. Flops are kept to similar values (1307 Mflops). After filter redistribution, most models surpass the base accuracy with less parameters. Results show average of three repetitions.\\}
\footnotesize
\begin{tabular}{l|rr|rr|rr|llll}
         & \multicolumn{2}{l|}{Param}      & \multicolumn{2}{l|}{Mem}  & \multicolumn{2}{l|}{Inference} &         &          &         & tiny     \\
Template & \multicolumn{2}{l|}{(Millions)} & \multicolumn{2}{l|}{(GB)} & \multicolumn{2}{l|}{Time (ms)} & cifar10 & cifar100 & cinic10 & Imagenet \\ \hline
base     & 23.52           & \% ↓          & 185.5        & \% ↓       & 5.38           & \% ↓          & 95.91± 0.29 & 78.31± 0.54 & 88.78± 0.93 & 65.57± 0.47 \\ \hline
a        & 14.17           & 39.7          & 146.8        & 20.8       & 4.62           & 14.1          & 96.10± 0.07 & 79.00± 0.05 & 89.60± 0.05 & 66.06± 0.53 \\
b        & 4.85            & 79.3          & 132.1        & 28.7       & 4.30           & 20.0          & 96.07± 0.08 & 78.91± 0.08 & 89.36± 0.09 & 65.01± 0.43 \\
c        & 3.48            & 85.2          & 128.9        & 30.5       & 4.28           & 20.4          & 96.13± 0.20 & 77.92± 0.18 & 89.27± 0.15 & 64.07± 0.17 \\
d        & 8.36            & 64.4          & 125.8        & 32.1       & 4.36           & 18.9          & 96.20± 0.11 & 79.43± 0.24 & 89.30± 0.29 & 65.59± 0.39 \\
e        & 3.68            & 84.3          & 132.1        & 28.7       & 4.31           & 19.8          & 95.79± 0.03 & 77.99± 0.48 & 89.15± 0.02 & 64.49± 0.57

\end{tabular}
\label{tab:templates_accuracy_resnet50}
\end{table*}

\begin{figure}
  \begin{multicols}{2}
    \includegraphics[width=0.95\linewidth]{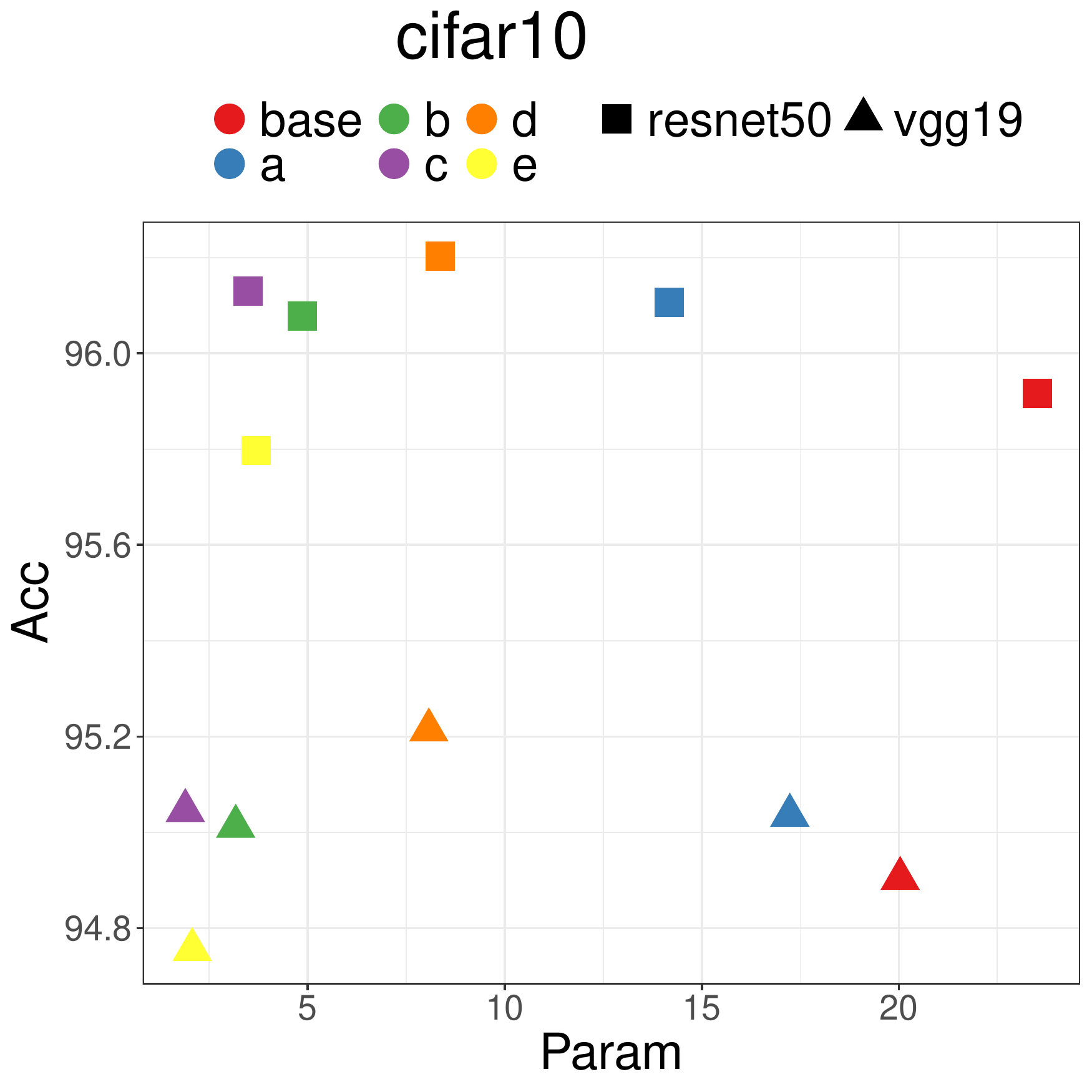}\par
    \includegraphics[width=0.95\linewidth]{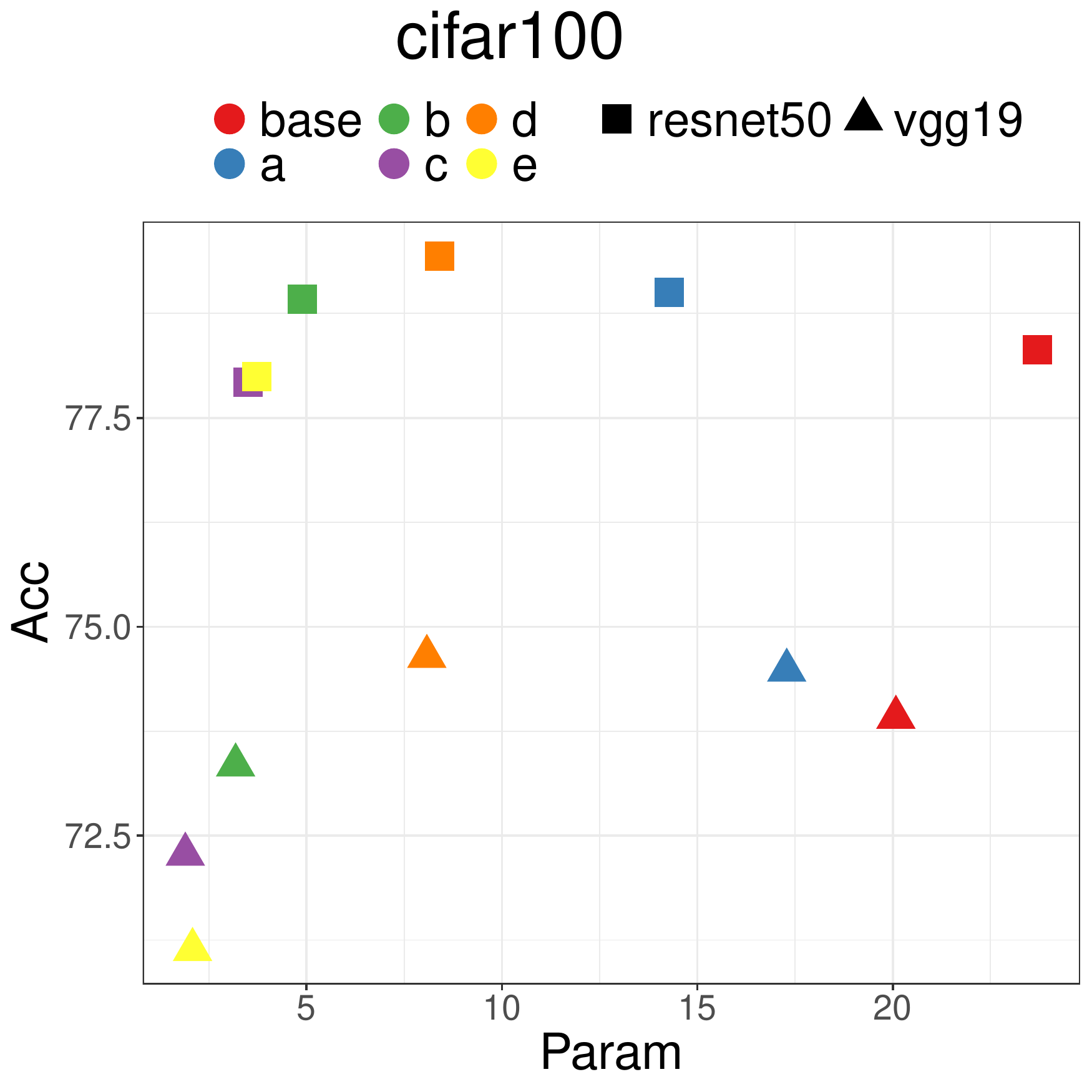}\par
  \end{multicols}
  \begin{multicols}{2}
    \includegraphics[width=0.95\linewidth]{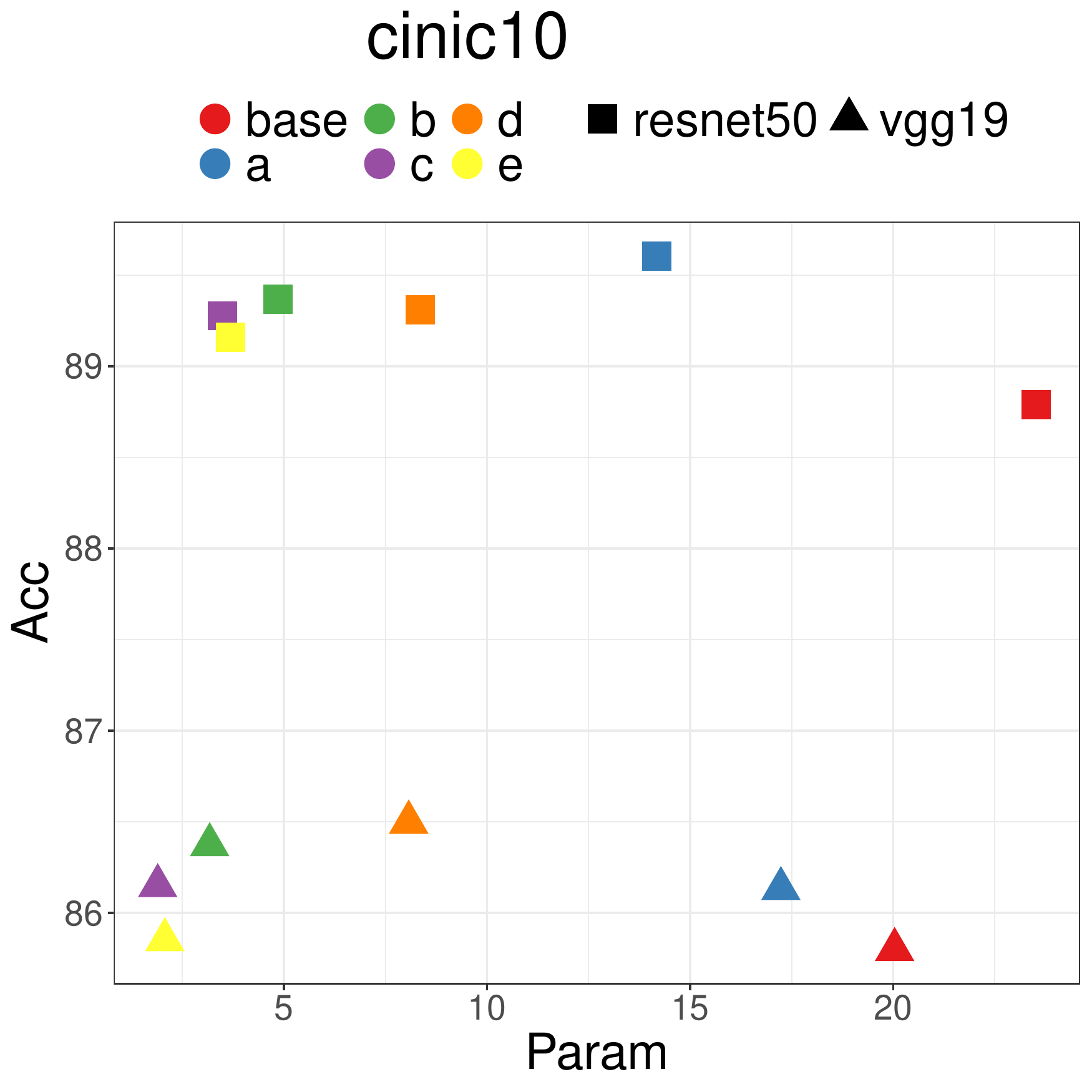}\par
    \includegraphics[width=0.95\linewidth]{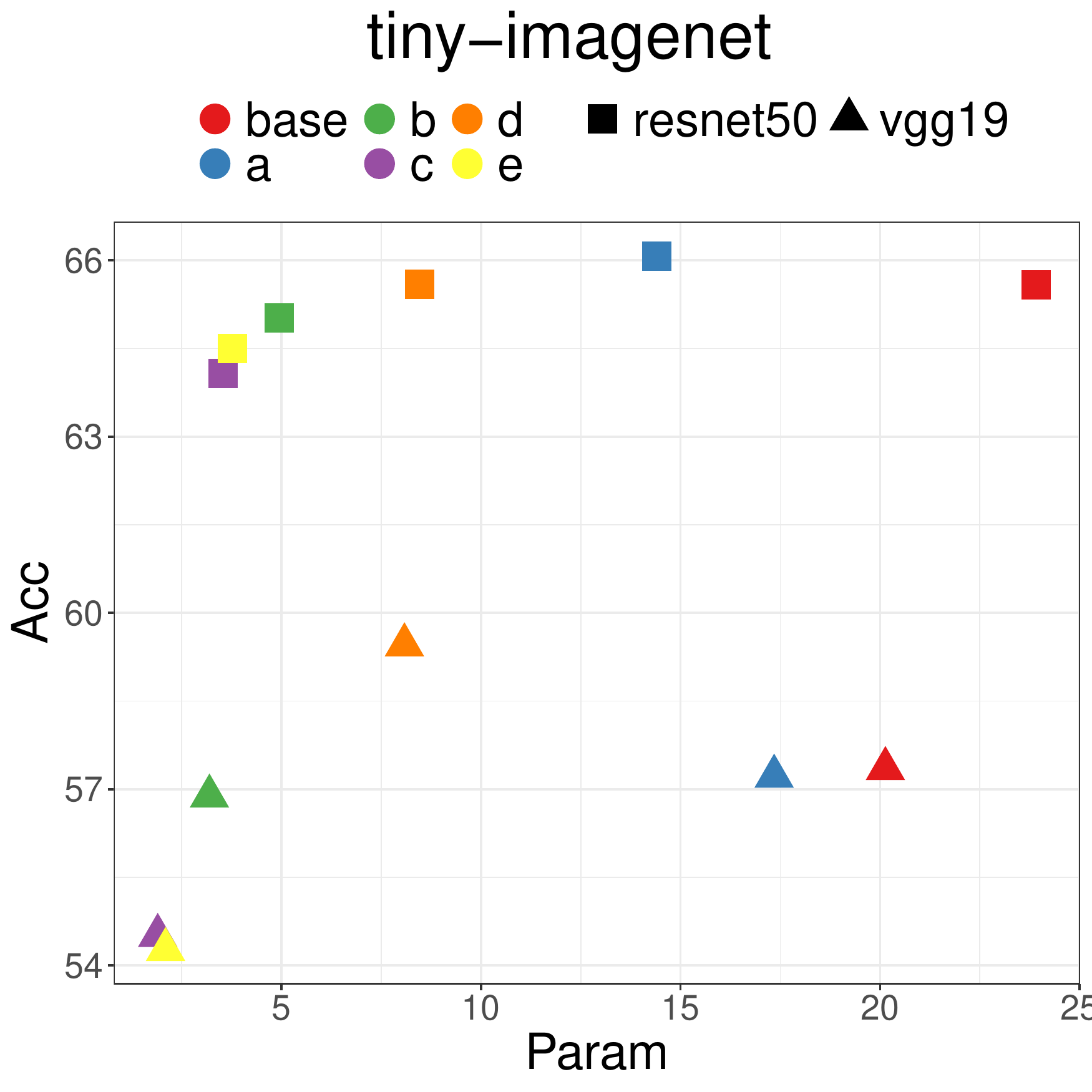}\par
  \end{multicols}
    \caption{Accuracy of models after applying templates reported for four datasets. Base is the original distribution of filters. In many cases, templates outperform the base architecture, however, all of them use much less parameters than the base model. Note that models produced with templates from VGG have less than a third of FLOPs of those from ResNet.}
\label{fig:models_accuracy}
\end{figure}

We found that there is a frequent behaviour related to each template that is clearly observed in figure \ref{fig:models_accuracy} . Accuracy improves in almost all datasets with template \textbf{d}. Template \textbf{b} emerges as a good trade off between resource consumption and accuracy and templates \textbf{c} and \textbf{e} give the biggest reduction in resources by sacrificing accuracy.

\section{Conclusions}\label{chap:conclusions}

We show that the popular pyramidal design for filter distribution in convolutional neural networks can be improved by modifying base architectures with a small set of filter distributions that we call templates. On the CIFAR, CINIC and Tiny-Imagenet datasets, our results suggest that the original pyramidal distribution is not necessarily the best option for obtaining the highest accuracy or  highest resource efficiency. Models with the same amount of FLOPs, but different distributions produced by our templates, show improved accuracy for at least one template in all evaluated models and tasks. In terms of resource consumption, templates can help obtain a competitive accuracy vs to original models but using much less resources with up to 90\% less parameters and a memory footprint up to 79\% smaller.

Our approach allows a model's architect to apply a set of templates for changing the number of filters originally assigned to each layer before training from scratch. This redesign can be easily achieved without any previous training process to select particular weights. In essence, the application of filter distribution templates offers an alternative approach to the iteration-intensive automatic architecture search and model pruning methods.

Our work offers insights to model designers, both automated and manual, to construct more efficient models by introducing the idea of new distributions of filters for neural network models.  We hope this work inspires the re-think of assumptions on model building and helps to gather data for better understanding the design process of model-task pairs.

\subsubsection*{Acknowledgments}
\AcknowledgmentsText

\clearpage

\bibliographystyle{ieee_fullname}
\bibliography{bibliography}

\end{document}